\documentclass[10pt,twocolumn,letterpaper]{article}
\usepackage[accsupp]{axessibility} 
\usepackage{wacv}              

\usepackage{graphicx}
\usepackage{amsmath}
\usepackage{amssymb}
\usepackage{booktabs}

\usepackage{multirow}
\usepackage{graphics}

\usepackage[pagebackref,breaklinks,colorlinks]{hyperref}

\usepackage[capitalize]{cleveref}
\crefname{section}{Sec.}{Secs.}
\Crefname{section}{Section}{Sections}
\Crefname{table}{Table}{Tables}
\crefname{table}{Tab.}{Tabs.}

\def\wacvPaperID{162} 
\def\confName{WACV}
\def\confYear{2024}

\begin{document}

\title{Augment the Pairs: Semantics-Preserving Image-Caption Pair Augmentation \\for Grounding-Based Vision and Language Models}

\author{Jingru Yi, Burak Uzkent, Oana Ignat, Zili Li, Amanmeet Garg, Xiang Yu, Linda Liu\\
Amazon Prime Video\\
}
\maketitle

\begin{abstract}
Grounding-based vision and language models have been successfully applied to low-level vision tasks, aiming to precisely locate objects referred in captions. The effectiveness of grounding representation learning heavily relies on the scale of the training dataset. Despite being a useful data enrichment strategy, data augmentation has received minimal attention in existing vision and language tasks as augmentation for image-caption pairs is non-trivial. In this study, we propose a robust phrase grounding model trained with text-conditioned and text-unconditioned data augmentations. Specifically, we apply text-conditioned color jittering and horizontal flipping to ensure semantic consistency between images and captions. To guarantee image-caption correspondence in the training samples, we modify the captions according to pre-defined keywords when applying horizontal flipping. Additionally, inspired by recent masked signal reconstruction, we propose to use pixel-level masking as a novel form of data augmentation. While we demonstrate our data augmentation method with MDETR framework, the proposed approach is applicable to common grounding-based vision and language tasks with other frameworks. Finally, we show that image encoder pretrained on large-scale image and language datasets (such as CLIP) can further improve the results. Through extensive experiments on three commonly applied datasets: Flickr30k, referring expressions and GQA, our method demonstrates advanced performance over the state-of-the-arts with various metrics. Code can be found in \url{https://github.com/amzn/augment-the-pairs-wacv2024}.
\end{abstract}

\section{Introduction}
Phrase grounding identifies objects in a scene based on the understanding of language. It requires the model to comprehend the visual context and relate object regions with sentences or phrases~\cite{plummer2015flickr30k,yu2018rethinking,yang2019fast}. Compared to conventional object detection, phrase grounding alleviates the bottleneck of fixed vocabulary and is able to generalize to unseen categories and attributes based on learning of nuance concepts of the free-form text~\cite{Kamath2021MDETRM,yao2022detclip,yao2023detclipv2}. As a result, a dataset that involves rich region-phrase and image-language correspondences is important for generalizable phrase grounding. For instance, MDETR~\cite{Kamath2021MDETRM} surpasses previous works with an effective pretraining datasets of 1.3M image-text pairs. On the other hand, GLIP~\cite{li2022grounded} demonstrates strong zero-shot and few-shot transferability by scaling up visual concepts with 27M grounding data. 
Aside from data scale, existing works have barely explored data augmentation in the phrase grounding task, despite the significant role that data augmentation plays in defining effective predictions across various tasks~\cite{chen2020simple,cubuk2019autoaugment,dwibedi2017cut}. 

\begin{figure*}[thb!]
  \centering
  \includegraphics[width=0.89\linewidth]{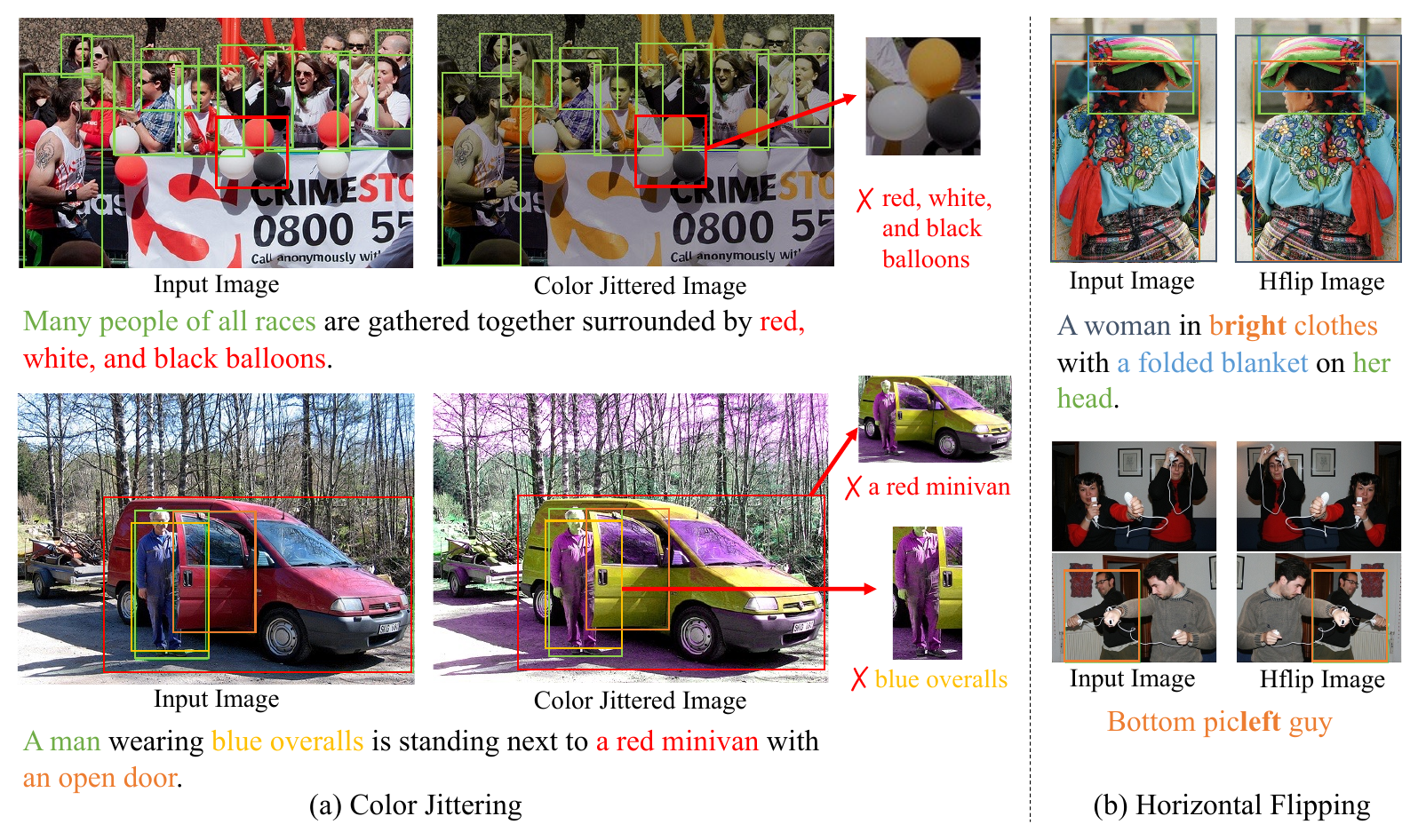}
  \caption{Examples of challenges in applying data augmentation in phrase grounding task. We highlight two common augmentations:  (a) color jittering and  (b) horizontal flipping. The bounding boxes in the image correspond to the same color phrases as those in the caption.}
  \label{fig:challenges} 
\end{figure*}

Data augmentation has been extensively studied and employed in object detection~\cite{girshick2015fast,carion2020end,dwibedi2017cut} to increase the density and variety of training samples~\cite{zoph2020learning}. For phrase grounding task, the sample shortage problem is more severe. For example, Flickr30k Entities~\cite{plummer2015flickr30k} contains 44.5k object categories, with only an average of 6.2 objects per category. Data augmentation could be crucial to improve model's generalization ability. To enhance phrase grounding understanding, some works~\cite{yao2022pevl,yang2023improving} employ data augmentations such as horizontal flipping in their pipeline.
However, applying data augmentation to phrase grounding can easily disrupt the image-language correspondence. For instance, as shown in Figure~\ref{fig:challenges} (a), color jittering can alter the colors of objects, causing a mismatch between the object regions and the corresponding noun phrases in the caption. Merely removing color-related words may lead to errors in ground-truth bounding boxes, as objects in different color may not be explicitly mentioned in the caption. Likewise, horizontal flipping is associated with words that convey left or right. A simple flip of words containing ``left'' or ``right'' can introduce image-caption misalignment (see Figure~\ref{fig:challenges}(b)).

To address these limitations, this paper proposes a novel text-conditioned augmentation approach, wherein we apply color jittering and horizontal flipping transformations to image-caption pairs that do not 
contain color-related keywords (e.g., red, yellow) and contain position-relevant words respectively. Furthermore, we utilize caption-independent data augmentations such as pixel-level and block-level masking to further enhance the learning of representations. In this paper, we pick one of the representative method MDETR, and apply augmentation strategies on top of it to demonstrate the effectiveness in improving the phrase grounding model on three 
datasets: Flickr30k~\cite{plummer2015flickr30k}, referring expressions~\cite{yu2016modeling} and GQA~\cite{hudson2019gqa}. These augmentation strategies can be seamlessly integrated into other orthogonal phrase grounding models by enriching image-caption correspondences in the training datasets. The experiments show that leveraging image encoder pretrained on larger-scale image-language datasets (e.g., CLIP~\cite{radford2021learning}) leads to additional performance gain. 
Our contributions can be summarized as: 
\begin{itemize}
    \item We propose text-conditioned 
    and text-unconditioned data augmentations to effectively enrich the data diversity, which can orthorgonally improve the vision and language grounding frameworks, e.g., MDETR.
    \item We show that by utilizing an image encoder pretrained on larger-scale vision and language datasets, such as CLIP, the embedding power can be significantly enhanced and thus improve the grounding performance.
    \item Extensive experiments on pretraining and downstream tasks demonstrate the superiority of the proposed method. Specifically, in the pretraining phrase grounding task, our method improves MDETR by 0.5\%, 3.3\% and 1\% AP on the validation set of Flickr30k, referring expressions and GQA datasets, respectively.
\end{itemize}

\section{Related Work}
\noindent \textbf{Grounding-based Vision and Language Models} 
We can categorize existing grounding-based image and language models into two categories: two-stage and single-stage. Two-stage methods~\cite{yu2018mattnet,chen2020uniter,lu2019vilbert} rely on off-the-shelf object detectors to get object proposals and then process the language query for the task of interest. On the other hand, single-stage methods~\cite{Kamath2021MDETRM,deng2021transvg,yang2020improving,chen2018real,li2022grounded,zhang2022glipv2,dou2022coarse,uzkent2023dynamic} avoid using a separate off-the-shelf object detector and perform end-to-end training for detecting the referred object, reducing the computational complexity of the two-stage methods. For example, MDETR~\cite{Kamath2021MDETRM} has trained an object detector (\textit{i.e.} DETR~\cite{carion2020end}) on a concatenation of learned image and language representations. Lite-MDETR~\cite{Lou_2022_CVPR} further reduces MDETR model size by leveraging a light-weight backbone and employing quantization.
The most recent vision and language models~\cite{Kamath2021MDETRM,deng2021transvg,li2021referring,li2022grounded,yang2022tubedetr} utilize large-scale transformers to improve the accuracy of the previous models with CNN backbones~\cite{chen2018real,yang2020improving}. Other recent works~\cite{radford2021learning,li2020unimo,shen2021much} including CLIP~\cite{radford2021learning} have developed image-language models trained on large-scale data with high-level image-to-text contrastive learning. We demonstrate that the phrase grounding model can be further improved by incorporating such models.

\begin{figure*}[thb!]
  \centering
  \includegraphics[width=0.85\linewidth]{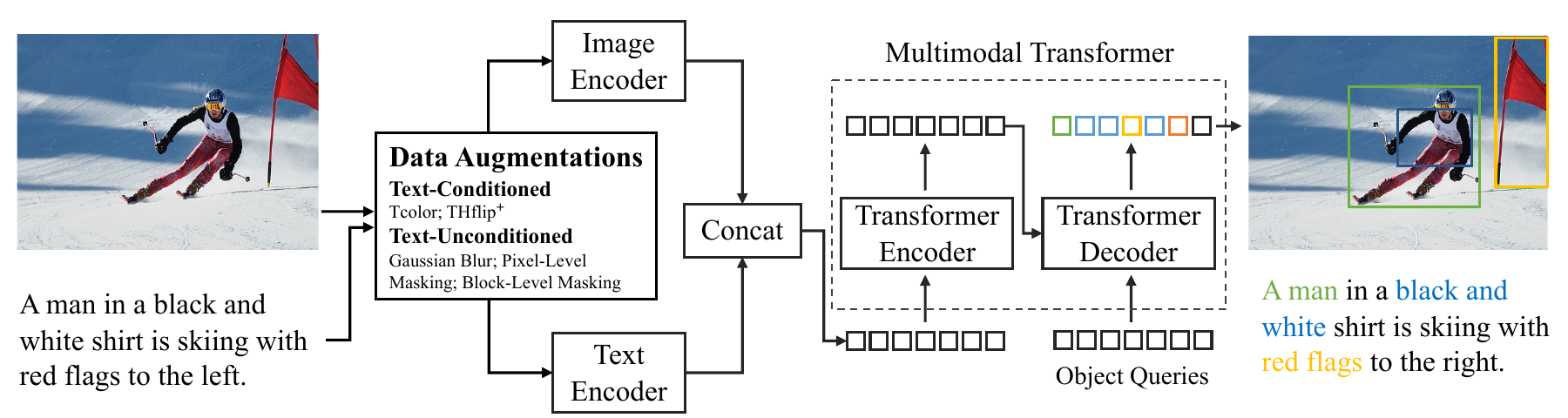}
  \caption{Illustration of phrase grounding framework, where input is an image-caption pair, and output is a set of grounded object regions mentioned in the caption.  During training, we apply 
  text-conditioned and text-unconditioned data augmentations to the input in order to increase sample's density and variety. 
  The representations from image and text encoders are concatenated and fed into a multimodal transformer, which learns to associate the textual and visual modalities for vision and language tasks.}
  \label{fig:architecture}
\end{figure*}

\begin{figure*}[thb!]
  \centering
  \includegraphics[width=0.94\linewidth]{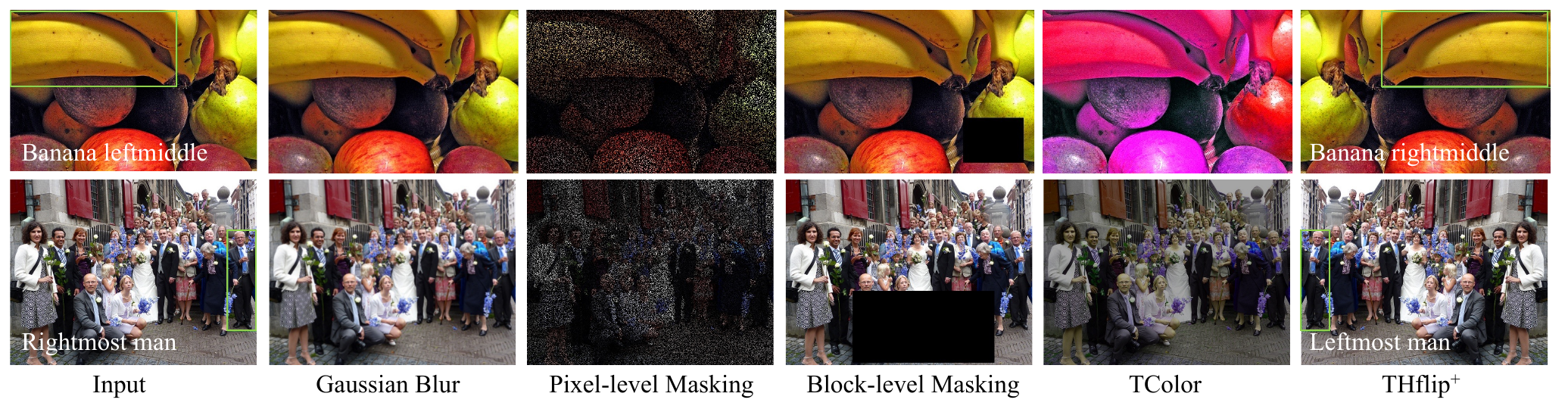}
  \caption{Illustration of text-conditioned and text-unconditioned data augmentations. Text-unconditioned data augmentations consist of Gaussian blur, pixel-level and block-level masking, while text-conditioned data augmentations involve text-conditioned color jitterring (TColor) and horizontal flipping (THflip$^+$). }
  \label{fig:augmetations}
\end{figure*}

\noindent \textbf{Data Augmentation} 
Data augmentation consistently leads to improved generalization~\cite{simard2002transformation,krizhevsky2017imagenet} in computer vision tasks. Elastic distortions with scale, translation and rotation, random cropping, image mirroring and color jittering are common data augmentations applied in classification models trained on natural images~\cite{simard2002transformation,krizhevsky2017imagenet,zhang2017mixup}. Compared to image classification, data augmentation is crucial for object detection as human annotations can be expensive and time-consuming~\cite{zoph2020learning}.  Image mirror and multi-scale training are the most widely used augmentation strategies for object detection~\cite{girshick2015fast,carion2020end}. Random erase~\cite{devries2017improved,zhong2020random}, additive noise~\cite{gilmer2019adversarial}, cut-and-paste~\cite{dwibedi2017cut}, augmentation policy learning~\cite{zoph2020learning} are also utilized in object detection to improve generalization performances for detection models. In this paper, we employed random erase and additive noise, and tailored the commonly used color jittering and image mirroring augmentations specifically for the phrase grounding task.


\section{Method}
\subsection{Preliminary From MDETR}
MDETR~\cite{Kamath2021MDETRM} model consists of an image and text encoders, and a multimodal transformer (see Figure~\ref{fig:architecture}). Given an image encoder $f_i$ and text encoder $f_t$ parameterized by $\theta_i$ and $\theta_t$, we denote the output representations $\textbf{z}_i \in \mathbb{R}^{N\times D}$ and $\textbf{z}_t \in \mathbb{R}^{M\times D}$ as
\begin{equation}
    \textbf{z}_i = f_i(x_i;\theta_i), \quad \textbf{z}_t = f_t(x_t;\theta_t),
\end{equation}
where $x_i$ and $x_t$ represent input image and text. $N$ and $M$ denote the number of image tokens and text tokens, respectively, $D$ is the feature dimension. The image encoder is parameterized by a CNN (\textit{e.g.}, ResNet~\cite{he2016deep}). The output image features are flattened as sequential image tokens, which are added with a sequence of position embeddings to preserve the spatial information. The text encoder learns text representations through a pre-trained transformer language model RoBERTa~\cite{liu2019roberta}. The image and text features are projected into a shared embedding space with a modality dependent linear projection. The modality-specific representations $\textbf{z}_i$ and $\textbf{z}_t$ are then concatenated and passed to a transformer encoder $f_e$, parameterized by $\theta_e$ as 
\begin{equation}
    \textbf{z}_e = f_e([\textbf{z}_i, \textbf{z}_t];\theta_e)
\end{equation}
where $\textbf{z}_e$ is the output of transformer encoder.  The output representation $\textbf{z}_e$ as well as object queries $\textbf{z}_q \in \mathbb{R}^{L\times D}$ are fed into a transformer decoder $f_d$, where $L$ is the number of object queries. We denote the output object embeddings  $\textbf{z}_o \in \mathbb{R}^{L\times D}$ of MDETR as
\begin{equation}
    \textbf{z}_o = f_d(\textbf{z}_e, \textbf{z}_q;|\theta_d),
\end{equation}
where $\theta_d$ is the transformer decoder parameters. 

\subsection{Data Augmentation}
As described in Figure~\ref{fig:challenges}, directly applying general data augmentations to phrase grounding task can lead to misalignment between image and caption. To overcome this challenge, we apply text-unconditioned and text-conditioned data augmentation to the input of our model.

\subsubsection{Text-Unconditioned Data Augmentation}
The text-unconditioned image augmentations are applied regardless of the text query. We carefully select Gaussian blur, pixel-level and block-level masking (see Figure~\ref{fig:augmetations}) in this direction. 


\noindent\textbf{Pixel-level and Block-level Masking.} Inspired by the recent works in masked signal modeling works~\cite{he2022masked,kwon2022masked}, we add pixel-level masking as noise to input images without 
reconstruction. In particular, we randomly mask the input image pixels with a probability $p$. We show that a simple pixel-level masking augmentation improves the phrase grounding performance significantly. In addition, block-level masking~\cite{zhong2020random} is adopted to randomly erase a block of pixels from input images (as shown in Figure~\ref{fig:augmetations}). The masking augmentations add occlusion to input images and force the model to learn to generalize well~\cite{zhong2020random} even though the present objects are not visually clear.

\subsubsection{Text-Conditioned Data Augmentation}
We further introduce text-conditioned data augmentations by modifying the input images in color and spatial space without breaking the image-caption correspondences of the phrase grounding task. Specifically, we introduce the novel text-conditioned color jittering and horizontal flipping.

\noindent\textbf{Text-Conditioned Color Jittering.} As shown in Figure~\ref{fig:challenges}(a), a general color jittering can bring errors when the caption contain color information. To address this limitation, we skip the color jittering when an input caption contains any color-related words, we term the method as TColor. In this way, we ensure the caption is still valid even though the image is altered in color space. 

\noindent\textbf{Text-Conditioned Horizontal Flipping.} Due to the complex interplay of the word combinations (see Figure~\ref{fig:challenges}(b)), simply replacing word containing ``left'' with ``right'' or vice versa when applying horizontal flipping in phrase grounding task would introduce errors. One way to address this limitation is to skip the horizontal flipping if the caption contains ``left'' or ``right''. We term this method as THflip. Although is still valid, THflip misses opportunities that could help the model learn the non-trivial connections between positional word and flipped image. To address this issue, we create a keyword list that contains words ``left'' or ``right'', and their variants with suffix ``-most, -side, -iest, -middle" or prefix such as ``upper-, top-, bottom-, far-", etc. For those captions that are not appearing in the keyword list, we choose not to conduct the horizontal flipping data augmentation.
%
We term the method as THflip$^+$. 
As depicted in the last column of Figure~\ref{fig:augmetations}, THflip$^+$ preserves image-caption consistency after modifying the caption. Through ablation experiments, we demonstrate that THflip$^+$ improves the  performance of phrase grounding task significantly especially on Referring expression comprehension dataset, when compared to THflip.



\subsection{Training Losses}
The training of our method involves three losses~\cite{Kamath2021MDETRM}: bounding box regression, soft token prediction and text-query contrastive alignment. We illustrate each loss below.

\noindent\textbf{Object-Text Contrastive Alignment.} 
To ensure the object representation is closer to the corresponding phrase text tokens in feature space compared to other objects, a contrastive loss (\textit{i.e.}, InfoNCE~\cite{oord2018representation}) is applied to the object embedding and text tokens. Given an object embedding $\textbf{z}_{oi}\in \mathbb{R}^{D}$ and its aligned text token set $T_i^{+}=\{\textbf{t} \in \mathbb{R}^{D}\}$ where $D$ is embedding length, the aligned contrastive loss for all objects can be formulated as:
\begin{equation}
    \mathcal{L}_{o}= \sum_{i=1}^L\frac{1}{|T_i^{+}|}\sum_{j\in T_i^{+}}-\log(\frac{\exp(\textbf{z}_{oi}^\top \textbf{t}_j/\tau)}{\sum_{k=1
    }^{E}\exp(\textbf{z}_{oi}^\top \textbf{t}_k/\tau)}),
\end{equation}
where $\tau=0.07$ is a temperature parameter~\cite{wu2018unsupervised}, $E$ is the maximum number of text tokens, $L$ is the number of object queries. Given a text token $t_j$ and its aligned object embedding set $O_j^+$, the contrastive loss for all text tokens is:
\begin{equation}
    \mathcal{L}_{t}= \sum_{j=1}^E\frac{1}{|O_j^{+}|}\sum_{i\in O_j^{+}}-\log(\frac{\exp(\textbf{t}_j^\top \textbf{z}_{oi}/\tau)}{\sum_{k=1
    }^{L}\exp(\textbf{t}_j^\top \textbf{z}_{ok}/\tau)}).
\end{equation}
The object-text contrastive loss can be expressed as $\mathcal{L}_{align}=(\mathcal{L}_{o}+\mathcal{L}_{t})/2$.

\begin{table*}[htb!]
    \centering
    \resizebox{0.8\linewidth}{!}{
    \begin{tabular}{c c c c c c c c c}
    \toprule
    \multirow{2}{*}{Method} & Image &  \multicolumn{2}{c}{Flickr30k} & \multicolumn{4}{c}{Referring Expressions} & \multicolumn{1}{c}{GQA}
    \\
   \cmidrule(l){3-4} \cmidrule(l){5-8} \cmidrule(l){9-9}
    & Encoder & AP & R@1 & AP & RefCOCO R@1 & RefCOCO+ R@1 & RefCOCOg R@1 & AP \\
    \midrule
    MDETR~\cite{Kamath2021MDETRM} & RN101  & 32.2 & 71.7 & 27.3 & 60.7 & 48.5 & 47.5 & 19.6 \\ 
    Ours& RN101  & 35.6&75.2 &32.6&66.4&54.9&51.8&23.1 \\ 
    Ours& RN101-CLIP &40.2&78.2&37.1&67.1&55.7&54.0&25.9\\ 
    Ours& ViT-B-CLIP &41.1&78.5&39.5&71.1&57.6&54.1&27.8\\
    \midrule
    MDETR~\cite{Kamath2021MDETRM}\dag & RN101  & 52.6 & 82.3  & 46.9 & 72.6 & 58.1 & 55.3 & 39.4 \\
    Ours\dag & RN101 & 53.1 & 83.3 & \textbf{50.2} & \textbf{74.8} & 61.0 & 57.1 & 40.4 \\
    MDETR~\cite{Kamath2021MDETRM}\dag & RN101-CLIP  & 54.0 & 83.5 & 45.9 & 71.2 & 57.1 & 54.4 & 40.9 \\
    Ours\dag & RN101-CLIP & \textbf{54.7} & \textbf{84.2} & 49.2 & 72.7 & \textbf{61.2} & \textbf{57.4} & \textbf{42.4}\\
    \bottomrule
    \end{tabular}}
    \caption{Phrase grounding evaluation results on validation sets of  Flickr30k~\cite{plummer2015flickr30k}, referring expressions~\cite{yu2016modeling} and GQA~\cite{hudson2019gqa}. Unless otherwise specified, models were trained on 256$\times$256 pixel images, while the input resolution of ViT-B was 224$\times$224. We denote CLIP for encoders pretrained from~\cite{radford2021learning}. Models with $\dag$ were trained on 800$\times$1333 pixel images. 
    }
    \label{tab:visual_grounding_results}
\end{table*}

\noindent\textbf{Soft Token Prediction.}
Following MDETR, rather than classifying the detected object, we utilize a soft token prediction method to identify the span of text tokens from input caption for each matched object. Given an object embedding $\textbf{z}_{oi}\in \mathbb{R}^D$ where $i$ indexes the predicted object, MDETR applies a linear layer to get the soft token predictions: $\textbf{s}_i = f(\textbf{z}_{oi})$, where $f:\mathbb{R}^D\rightarrow\mathbb{R}^E$ is a linear transformation function, $E$ is the maximum number of text tokens. Cross entropy loss is utilized to train the soft token predictions:
\begin{equation}
    \mathcal{L}_{stoken} = -\frac{1}{n^+}\sum_{i=1}^L\sum_{j=1}^{E}s_{ij}^*\log \frac{\exp (s_{ij})}{\sum_{k=1}^{E}\exp(s_{ik})},
\end{equation}
where $\textbf{s}_i^*$ is an uniform distribution of all positive tokens~\cite{Kamath2021MDETRM}, $n^+$ is the total number of matched objects, $L$ is the number of object queries.

\noindent\textbf{Bounding Box Regression.}
For bounding box coordinates regression, a multi-layer perceptron (MLP) module is applied to the object embedding $\textbf{z}_{oi}\in \mathbb{R}^D, i=1,\cdots,L$. Suppose the predicted box coordinates are $\textbf{c}_i\in\mathbb{R}^4$, the bounding box regression loss is devised as:
\begin{equation}
    \mathcal{L}_{box}=\frac{1}{n^+}\sum_i^L (L_1(\textbf{c}_i, \textbf{c}^*_i)+(1-\text{GIoU}(\textbf{c}_i, \textbf{c}^*_i))),
\end{equation}
where $c^*$ is the ground-truth box coordinates for matched objects, $n^+$ indicates the total number of matched objects. The bounding box regression loss combines $L_1$ loss and generalized intersection over union (GIoU)~\cite{rezatofighi2019generalized} loss.

\section{Experimental Details}
\noindent\textbf{Pretraining Datasets.}  We follow the same pretraining strategy as MDETR~\cite{Kamath2021MDETRM}.
Specifically, the training images are created from a combination of MSCOCO~\cite{lin2014microsoft}, Flickr30k~\cite{plummer2015flickr30k}, and Visual Genome (VG)~\cite{krishna2017visual}, where annotations are merged from Flickr entities~\cite{plummer2015flickr30k}, VG regions~\cite{krishna2017visual}, referring expressions \cite{yu2016modeling,mao2016generation} and GQA train balanced set~\cite{hudson2019gqa}. 
The dataset comprises bounding box annotations for objects mentioned in the language query, including 200k images and 1.3M aligned image-caption pairs. Among the annotations, Flickr30k Entities~\cite{plummer2015flickr30k} contains 31.8k images with 
5 sentences per image. It involves 44.5k object categories with 6.2 objects per category.  MSCOCO~\cite{lin2014microsoft} contains 37k images where annotations are collected from referring expressions (RefCOCO \cite{yu2016modeling}, RefCOCO+ \cite{yu2016modeling} and RefCOCOg \cite{mao2016generation}). 
Visual Genome~\cite{krishna2017visual} contains 108k image with 18k object categories where each image has 50 descriptions. We present the phrase grounding performance results for the pretraining task on the validation sets of Flickr30k~\cite{plummer2015flickr30k}, referring expressions~\cite{yu2016modeling} and GQA~\cite{hudson2019gqa}.


\noindent\textbf{Evaluation Metrics.} 
We use average precision (AP) and Recall@K (R@K)) as
evaluation metrics. AP~\cite{ren2015faster} is a standard evaluation metric for object detection, it is the average areas under Precision-Recall curve at IoU threshold ranges from 0.5 to 1 with an interval of 0.05. R@K~\cite{plummer2015flickr30k} measures the percentage of queries for which a correct match has rank of at most K.

\noindent\textbf{Implementation Details} The pretraining stage takes 40 epochs on V100 GPUs with an effective batch size of 128 for models with 256$\times$256 pixel images, and a batch size of 64 for 800$\times$1333 pixel images. We use the same implementation as MDETR. In particular, 
We use AdamW~\cite{loshchilov2017decoupled} as optimizer. Learning rate is initialized as 1e-4, which is divided by 10 at epoch 30. We apply the proposed augmentations randomly on input image-caption pairs. 


\begin{figure*}[thb!]
  \centering
  \includegraphics[width=0.85\linewidth]{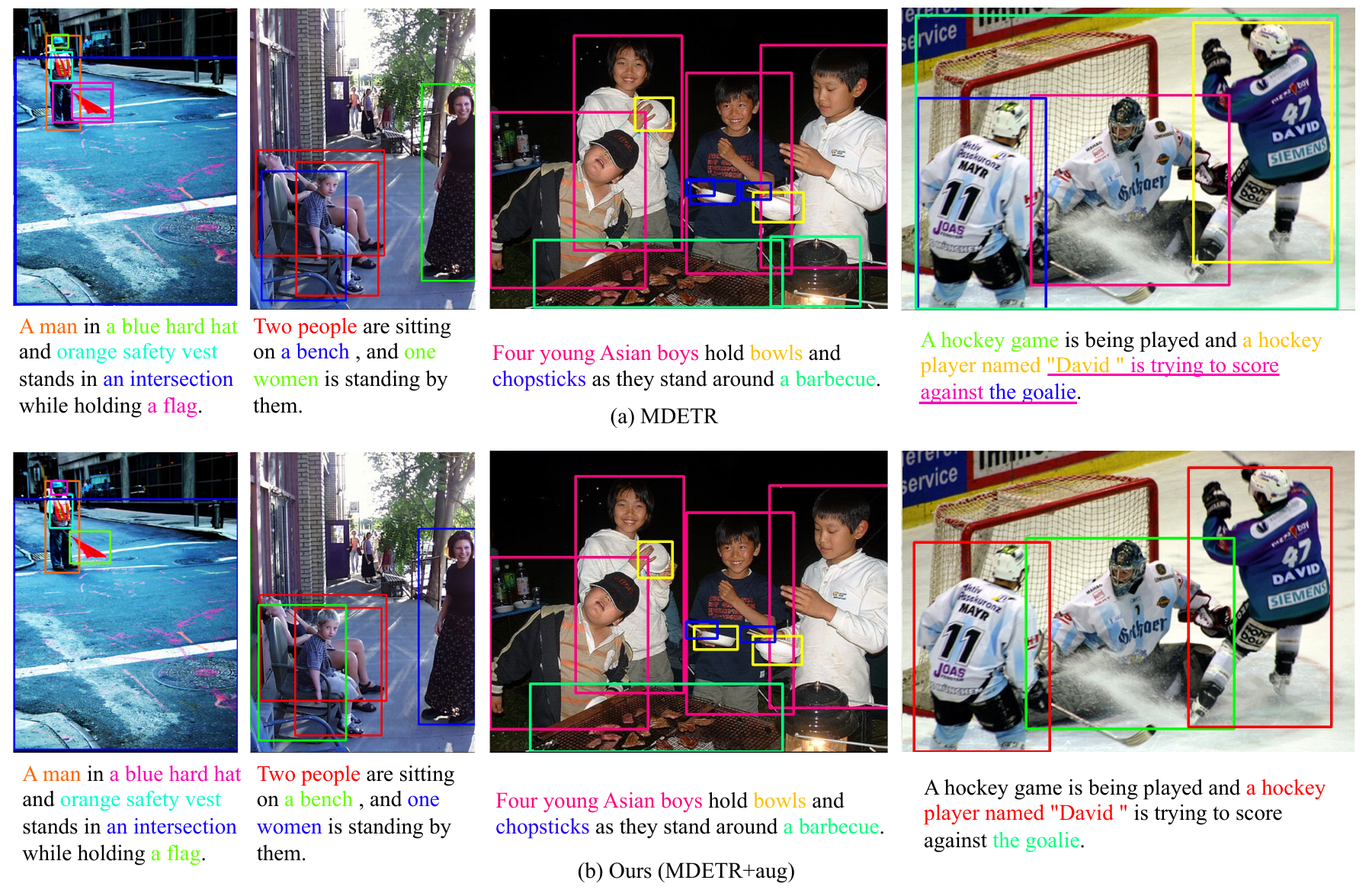}
  \caption{Visualization of phrase grounding prediction results with 800$\times$1333 pixel images on Flickr30k validation datasets. The bounding boxes with different color correspond to the phrase with the same color in the caption. Underscore distinguishes overlapped phrases.}
  \label{fig:visualization}
\end{figure*}

\begin{figure*}[thb!]
  \centering
  \includegraphics[width=0.85\linewidth]{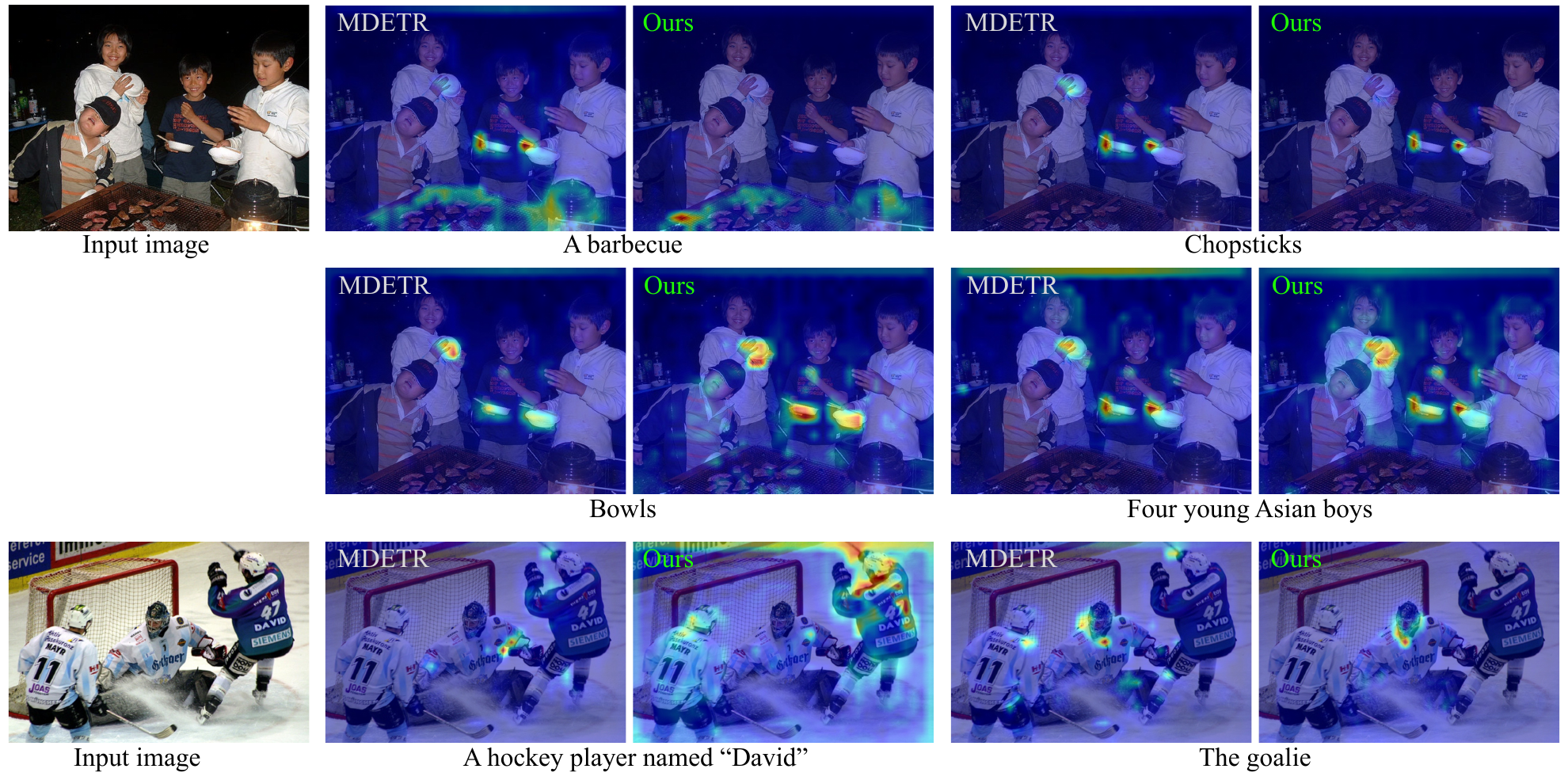}
  \caption{Visualization of attention maps queried by phrases.}
  \label{fig:attention}
\end{figure*}

\section{Results And Discussions}
In this section, we firstly show our method on pretraining phrase grounding task. In particular, we pick one of the state-of-the-art frameworks, \textit{i.e.}, MDETR~\cite{Kamath2021MDETRM}), as our baseline to demonstrate the effectiveness of the proposed data augmentation. Next, we evaluate our model on two downstream tasks: phrase grounding and referring expression comprehension. Finally, a detailed ablation is presented to highlight different augmentation modules.

\begin{table*}[th]
    \centering
    \resizebox{0.85\linewidth}{!}{
    \begin{tabular}{ c c c c c c c c c c c}
    \toprule
    \multirow{2}{*}{Method}
    &  \multirow{2}{*}{Pre-training data}
     & \multicolumn{3}{c}{RefCOCO} & \multicolumn{3}{c}{RefCOCO+} &
     \multicolumn{2}{c}{RefCOCOg}
     \\ 
    \cmidrule{3-11}
    & & val & testA & testB & val & testA & testB & val & test \\
    \midrule
    MAttNet\cite{Yu2018MAttNetMA}  &None & 76.65&  81.14&  69.99&  65.33&  71.62&  56.02&  66.58& 67.27 \\ 
    ViLBERT\cite{Lu2019ViLBERTPT} & CC (3.3M) & - & - & - & 72.34 & 78.52 & 62.61 & - & - \\
    VL-BERT$_L$\cite{Su2019VLBERTPO} & CC (3.3M)& - & - & - & 72.59 & 78.57 & 62.30 & - & - \\
    UNITER$_L$\cite{Chen2019UNITERLU} & CC, SBU, COCO, VG (4.6M) & 81.41 & 87.04 &74.17 & 75.90 & 81.45 & 66.70 & 74.86 & 75.77\\
    VILLA$_L$\cite{Gan2020LargeScaleAT} & CC, SBU, COCO, VG (4.6M) & 82.39 & 87.48 & 74.84 & 76.17 & 81.54 & 66.84 & 76.18 & 76.71  \\
    ERNIE-ViL$_L$\cite{Yu2020ERNIEViLKE} & CC, SBU (4.3M)& - & - & - & 75.95 & 82.07 & 66.88 & - & -\\
    MDETR-R101\cite{Kamath2021MDETRM} & COCO, VG, Flickr (200k)&86.75 & 89.58 & 81.41 & 79.52 & 84.09 & 70.62 & 81.64 & 80.89\\
    \midrule
    Ours-R101 & COCO, VG, Flickr (200k) & 87.47 & 90.24 & 81.83 & 79.91 & 84.49 &  71.18& 82.64 & 81.66 &  \\ 
    MDETR-R101-CLIP & COCO, VG, Flickr (200k) & 87.35 & 90.46 & 81.93 & 79.73 & 84.22 & 70.94 & 82.35 & 81.59 &  \\ 
    Ours-R101-CLIP & COCO, VG, Flickr (200k) & \textbf{87.72} & \textbf{90.60} & \textbf{82.20} & \textbf{80.45} & \textbf{85.01} & \textbf{71.50} &  \textbf{82.91} & \textbf{82.08} &  \\ 
    \bottomrule
    \end{tabular}}
    \caption{Results (R@1) on referring expression comprehension task.}
    \label{tab:RefCOCO-downstream}
\end{table*}

\subsection{Phrase Grounding Pretraining}
\noindent\textbf{With proposed Data Augmentation, our method consistently outperforms MDETR~\cite{Kamath2021MDETRM}.} In Table~\ref{tab:visual_grounding_results}, we evaluate the pretraining performance of our model on validation sets of Flickr30k~\cite{plummer2015flickr30k}, referring expressions~\cite{yu2016modeling} and GQA~\cite{hudson2019gqa}. In particular, on 256$\times$256 pixel images, the AP of our method exceeds MDETR by 3.4\%, 4.3\% and 3.5\% on Flickr30k, referring expressions and GQA datasets respectively. With ResNet101 pre-trained on CLIP, our method further improves 4.6\%, 4.5\% and 2.8\% AP on the three datasets. By replacing ResNet101 with vision transformer (\textit{i.e.,} ViT-B~\cite{dosovitskiy2020image}) as image encoder, our method obtains further improvement by 0.9\%, 2.4\% and 1.9\% AP. On the other hand, on 800$\times$1333 pixel images, our model surpasses MDETR by 0.5\%, 3.3\% and 1\% AP and by 0.7\%, 3.3\%, 1.5\% AP on the three datasets with RN101 and RN101-CLIP image encoder, respectively. 
Since the keyword list for THflip$^+$ is primarily derived from referring expressions, the significant improvement observed on this dataset suggests that introducing variability in image-caption associations enhances the model's learning ability for vision and language tasks.

\noindent\textbf{With proposed Data Augmentation, qualitative results exhibit better semantic understanding.} Figure~\ref{fig:visualization} shows qualitative results of phrase grounding on the Flickr30k validation dataset, using RN101 as backbone and 800$\times$1333 as input resolution. 
Our method shows increased robustness in suppressing redundant detection. For example, it effectively improves the redundant detection of ``a flag'' in the first image. In addition, it rectifies the erroneous detection of ``barbecue'' and ``chopsticks'' in the third column of images. More importantly, with the richer image-text correspondences introduced by data augmentation in the training dataset, the model shows better understanding of the context. As depicted in the last column of Figure~\ref{fig:visualization}, our method has correctly differentiated between ``the goalie'' and ``a hockey player'', whereas MDETR fails to recognize the ``goalie''. Moreover, by analyzing attention maps in Figure~\ref{fig:attention}, we observe that our model mainly relies on helmet features of the goalie to make decisions, whereas MDETR is influenced by ambiguous features. 

\begin{table}[t]
    \centering
    \resizebox{\linewidth}{!}{
    \begin{tabular}{c c c c c c c}
    \toprule
    \multirow{2}{*}{Method}
     & \multicolumn{3}{c}{Val} & \multicolumn{3}{c}{Test} \\ 
    & R@1 & R@5 & R@10 & R@1 & R@5 & R@10 \\
    \midrule
    BAN\cite{Kim2018BilinearAN} & - & - & - & 69.7 & 84.2 & 86.4\\
    VisualBert\cite{Li2019VisualBERTAS} & 68.1 & 84.0 & 86.2 & - & - & - \\
    VisualBert\dag\cite{Li2019VisualBERTAS} & 70.4 & 84.5 & 86.3 & 71.3 & 85.0 & 86.5 \\
    MDETR-RN101\cite{Kamath2021MDETRM} & 78.9 & 88.8 & 90.8 & - & - & -\\
    MDETR-RN101\dag*\cite{Kamath2021MDETRM} & 82.5 & 92.9 & 94.9 & 83.4 & 93.5 & 95.3 \\
    \midrule
   Ours-RN101\dag* & 83.3 & 93.0 & 95.1 & 83.7 & 93.6 & 95.4  \\
   MDETR-RN101-CLIP\dag* & 83.5 & 93.5 & 95.3 & 84.0 & 94.0 & 95.9 \\
   Ours-RN101-CLIP\dag* & \textbf{84.2}	& \textbf{93.7}	& \textbf{95.5}	& \textbf{84.7}	& \textbf{94.3}	& \textbf{95.9} \\

    \bottomrule
    \end{tabular}}
    \caption{Results of phrase grounding task on Flickr30k entities dataset (Any-Box protocol~\cite{Kamath2021MDETRM}). Models with $\dag$ are pre-trained on COCO, models with $*$ are also pre-trained on VG and Flickr 30k.}
    \label{tab:flickr30k-downstream}
\end{table}


\noindent\textbf{Failure Cases.} We also visualize some failure cases in Figure~\ref{fig:visualization}. For example, in the third column, both models fail to identify the chopsticks held by the second boy. As the color of the chopsticks is similar to the boy's shirt, this may indicate that the framework has limitations in detection of small objects with a similar color as the background. In the last image, our model identifies two persons as ``a hockey player named David'', indicating that the model is incapable of recognizing text ``David'' on the shirt. In Figure~\ref{fig:attention}, we also notice that MDETR does not consider text ``David'' even though it localizes the correct person. This may highlight the importance of text recognition from visual features for phrase grounding, which remains challenging due to limited training data. 
Nevertheless, the visualization results reveal that augmentations introduced in this work have significantly enhanced the model's robustness in the phrase grounding task.

\begin{table*}[htb!]
    \centering
    \resizebox{0.93\linewidth}{!}{
    \begin{tabular}{c c c c c | c c |c c c c| c}
    \toprule
    Gaussian & \multirow{2}{*}{THflip} & \multirow{2}{*}{TColor} & Pixel & Block & \multicolumn{2}{c|}{Flickr30k} & \multicolumn{4}{c|}{Referring Expressions} & GQA \\
    Blur & &  & Mask & Mask & AP & R@1 & AP & RefCOCO R@1 &  RefCOCO+ R@1 & RefCOCOg R@1 & AP \\
    \midrule
    &&&&&32.2 & 59.0&27.3&60.7&48.5&47.5&19.6\\
     \checkmark &&&&&32.8 &60.5&27.1&59.6&47.5&47.5&19.0\\ 
     &\checkmark &&&&34.0&74.3&28.3&59.7&50.6&49.2&21.7\\
     &\checkmark$^+$& &&&34.8&74.2&\textbf{34.0}&\textbf{69.9}&54.0&\textbf{53.4}&21.4\\
     &&\checkmark &&&32.4&72.1&29.6&63.9&49.8&48.4&20.2\\
     &&&\checkmark &&34.5&73.1&29.8&63.9&50.3&48.9&20.6\\
     &&&&\checkmark &33.8&73.6&29.7&65.4&50.7&49.7&21.1\\
      &\checkmark&\checkmark&\checkmark&\checkmark&35.8&75.2&31.2&65.3&52.6&50.5&\textbf{23.5} \\
     \checkmark&\checkmark&\checkmark&\checkmark&\checkmark&\textbf{36.3}&74.5&31.4&64.3&51.5&52.3&23.1 \\ 
     \checkmark&\checkmark$^+$&\checkmark&\checkmark&\checkmark&35.6&\textbf{75.3}&32.6&66.4&\textbf{54.9}&51.8&23.1 \\
    \bottomrule
    \end{tabular}}
    \caption{Ablation studies with different augmentation strategies on pretraining task. We evaluate the model performances on validation set of Flickr30k~\cite{plummer2015flickr30k}, referring expressions~\cite{yu2016modeling} and GQA~\cite{hudson2019gqa}. THflip and TColor refer to text-conditioned horizontal flipping and color jittering. Symbol $^+$ indicates THflip$^+$. Note that we use 224$\times$224 pixel images for this experiments. 
    }
    \label{tab:ablation_aug}
    \vspace{-2mm}
\end{table*}

\subsection{Downstream Tasks}
We finetune our model from pretraining stage for two downstream tasks: phrase grounding and referring expression comprehension, using the same training setting as MDETR~\cite{Kamath2021MDETRM}. Notice that there is no additional data augmentation applied for downstream tasks finetuning.

\noindent\textbf{Phrase Grounding}  The phrase grounding downstream task is performed on Flickr30k entities dataset~\cite{plummer2015flickr30k}. 
We follow the same setup as in MDETR~\cite{Kamath2021MDETRM} by taking the top 100 bounding box predictions and soft token predictions to align the bounding boxes to the caption. We compare our method to the prior works under the ANY-BOX protocol~\cite{Kamath2021MDETRM}. Similar to MDETR, the evaluation is conducted on models from pretraining stage, as additional fine-tuning does not yield further performance improvement. The results  presented in  Table~\ref{tab:flickr30k-downstream} demonstrate that the proposed method consistently outperforms MDETR and other state-of-the-art methods. By further leveraging a backbone pretrained on large-scale vision and language datasets (\textit{i.e.}, CLIP~\cite{radford2021learning}), our model exhibits even more superior performance. This evidences that larger-scale foundation model enables better representation power benefited from the large language models.

\noindent\textbf{Referring Expression Comprehension} 
Referring expression comprehension is a task that localizes the objects being referred from input image. In this task, only one box will be returned for each input expression. We present the results on three commonly applied datasets, RefCOCO \cite{yu2016modeling}, RefCOCO+ \cite{yu2016modeling} and RefCOCOg \cite{mao2016generation}. Following the same setting as MDETR, we finetune our model for 5 epochs as it is slightly different from pretraining phrase grounding task. As shown in Table~\ref{tab:RefCOCO-downstream}, with the same backbone, our method consistently improves the performance over MDETR on both the validation and test splits across the three datasets, suggesting that the proposed data augmentations effectively enhance the learned vision and language representation. 

\subsection{Ablation Study} 
To study the effectiveness of the proposed data augmentations, we ablate our model on 256$\times$256 pixel images and report the model performances of pretraining phrase grounding task in Table~\ref{tab:ablation_aug}. We observe that the Gaussian blur alone is not effective on referring expressions and GQA datasets. 
On referring expressions, THflip$^+$ outperforms THflip by 5.7\% in terms of AP. 
It suggests that augmenting both images and captions can effectively enrich the data variance and thus improve the learned embedding representation power. TColor improves the baseline on three datasets, but it is not as effective as THflip$^+$ and pixel and block-level masking. This is attributed to the fact that TColor skips augmentation when a caption contains color-related words ($\sim$ 49\% in pretraining datasets), thus limiting the variance it could bring to the training samples.
Pixel-level or block-level masking introduce further difficulties for the model to detect the occluded objects and therefore it is an effective method in improving the representations. Among all the augmentations, THFlip$^+$ achieves the best on referring expressions. By combining all the augmentations, our approach further improves the performance of THflip$^+$ on Flickr30k and GQA datasets without much sacrifice on referring expressions.

\begin{figure}[t]
  \centering
  \includegraphics[width=\linewidth]{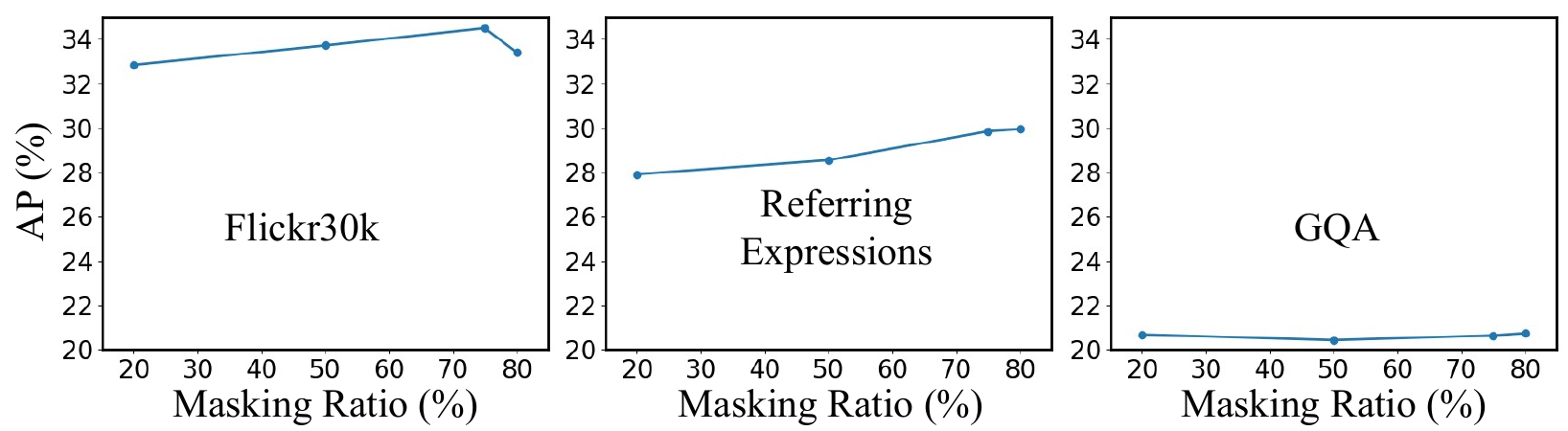}
  \caption{Phrase grounding performances with pixel-level masking augmentation only at different masking ratio on three datasets.}
  \label{fig:masking_ratio_ablation}
  \vspace{-3mm}
\end{figure}

\noindent\textbf{Masking Ratio.} To study the influence of masking ratio of pixel-level masking augmentation on phrase grounding task, we visualize the results at masking ratio of 20\%, 50\%, 75\% and 80\% on three datasets in Figure~\ref{fig:masking_ratio_ablation}. On Flickr30k, the masking ratio shows similar trend as MAE~\cite{He2021MaskedAA}, where 75\% ratio achieves the best performance. The higher masking ratio on referring expressions shows better performance, while the AP value is similar at 75\% and 80\% masking ratio. On GQA dataset, the AP value has minor difference across different masking ratios. Therefore, we empirically select 75\% ratio for the pixel-level masking augmentation.

\section{Conclusion} 
In this paper, we address the challenges that traditional data augmentations can hardly reserve the semantics in grounding-based vision and language tasks, as they can disrupt the image-caption correspondences. We propose text-conditioned (TColor and THflip$^+$) and text-unconditioned (pixel-level and block-level masking) data augmentations to enrich the image-caption density and diversity while preserving semantic coherence between object regions and corresponding phrases. Achieving this is challenging due to the complex interplay of word combinations.
With extensive experiments, we demonstrate that our method can effectively enhance the learned feature representations for grounding-based vision and language tasks.  
Further ablations show the effectiveness of our proposed augmentations against traditional Gaussian blur and masking operations. Future work will focus on how to generalize to an even broader range of the augmentations to further expand the variation of the input space.

\medskip

\end{document}